\documentclass{article}

\usepackage{PRIMEarxiv}

\usepackage[utf8]{inputenc} 
\usepackage[T1]{fontenc}    
\usepackage{hyperref}       
\usepackage{url}            
\usepackage{booktabs}       
\usepackage{amsfonts}       
\usepackage{nicefrac}       
\usepackage{microtype}      
\usepackage{lipsum}
\usepackage{fancyhdr}       
\usepackage{graphicx}       
\usepackage{multirow}
\usepackage{tabularx,array} 
\usepackage{pifont}
\usepackage{xcolor}

\graphicspath{{media/}}     

\newcommand{\cmark}{\ding{51}}%
\newcommand{\xmark}{\textcolor{red}{\ding{55}}}%
\newcommand{\hflogo}{\raisebox{-0.2ex}{\includegraphics[height=1.2em]{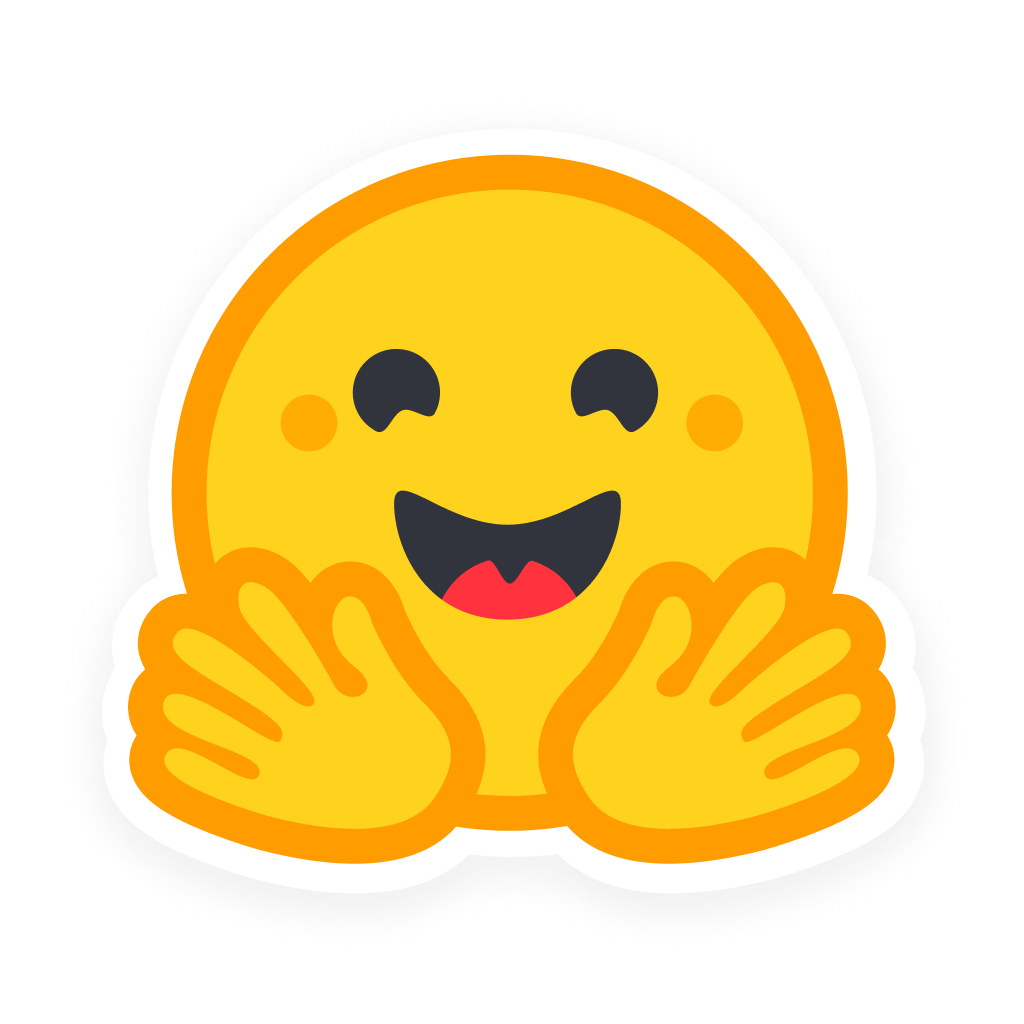}}}

\title{PubMed-OCR: PMC Open Access OCR Annotations}

\author{
  Hunter Heidenreich, Yosheb Getachew, Olivia Dinica, Ben Elliott \\
  Roots.ai \\ 
  \texttt{ai-ml@roots.ai} \\
  \\
  \href{https://huggingface.co/datasets/rootsautomation/pubmed-ocr}{\hflogo\ \textbf{Dataset:}\  \texttt{huggingface.co/datasets/rootsautomation/pubmed-ocr}}
}

\begin{document}

\maketitle

\begin{abstract}
\emph{PubMed\mbox{-}OCR} is an OCR-centric corpus of scientific articles derived from PubMed Central Open Access PDFs. Each page image is annotated with Google Cloud Vision and released in a compact JSON schema with word-, line-, and paragraph-level bounding boxes. The corpus spans 209.5K articles (1.5M pages; ~1.3B words) and supports layout-aware modeling, coordinate-grounded QA, and evaluation of OCR-dependent pipelines. We analyze corpus characteristics (e.g., journal coverage and detected layout features) and discuss limitations, including reliance on a single OCR engine and heuristic line reconstruction. We release the data and schema to facilitate downstream research and invite extensions.
\end{abstract}

\section{Introduction}

PDFs and scanned documents are ubiquitous in business, government, education, research, and healthcare. 
To realize AI copilots that meaningfully reduce repetitive work, systems must robustly understand real-world documents~\cite{wang2025document}.

Open data is central to this goal. 
By democratizing access and standardizing formats, open corpora enable better models and algorithms~\cite{kapoor2024societal}, support reproducibility~\cite{pineau2021improving}, broaden participation, and yield stronger benchmarks. 
This dynamic is evident in broad-spectrum LLM training sets~\cite{gao2020pile,laurenccon2022bigscience,kocetkov2022stack,soldaini2024dolma,weber2024redpajama} and in large-scale multimodal resources~\cite{schuhmann2021laion,schuhmann2022laion,gu2022wukong}. 
In document processing, open datasets have repeatedly shaped progress, from \emph{IIT\mbox{-}CDIP}~\cite{lewis2006building,soboroff_cdip_2022} and its derivatives~\cite{harley2015evaluation,jaume2019funsd,zhu2007automatic,agam2006cdip,ucsf2007ltdl,mathew2021docvqa} to PubMed-derived resources that target complex scientific articles~\cite{tkaczyk2012grotoap,tkaczyk2014grotoap2,zhong2019publaynet,smock2022pubtables}).

A pattern of co-reinforcement is evident: open data leads to stronger models, and stronger models catalyze the creation of further open data. 
While closed-source systems are often scaled further than open models, their outputs can be released to accelerate community progress, as seen with \emph{OCR\mbox{-}IDL}~\cite{biten2022ocr}. 
Open models are also combined in ensembles to improve data quality for state-of-the-art systems.
For example, DeepSeek\mbox{-}OCR~\cite{wei2025deepseek} scaled supervision by leveraging PP\mbox{-}DocLayout~\cite{sun2025pp}, MinerU~\cite{wang2024mineru}, GOT\mbox{-}OCR2.0~\cite{wei2024general}, and PaddleOCR~\cite{cui2025paddleocr30technicalreport}. 

These effects extend beyond document processing.
Advances in OCR yield improved LLM training corpora~\cite{kydlicek2025finepdfs}, since OCR provides the translation layer from optical signals to discrete text. 
Without faithful translation, information remains effectively invisible to text-only models. 

\paragraph{This work.}
We introduce \emph{PubMed\mbox{-}OCR}, built from the same open-access database used by prior PubMed resources. 
Unlike approaches that align text or regions mined from digital PDFs to JATS XML (a process prone to parser noise, heuristic dependencies, and missed text from scanned documents), we annotate page images directly with a commercial OCR system (Google Vision OCR) to produce word-, line-, and paragraph-level supervision. 
We provide corpus statistics and qualitative examples, and release the resource to support model development, benchmark curation, and related research.

\section{Related Work}

\subsection{PMCOA-Derived Layout and Table Datasets}

PubMed Central Open Access (PMCOA) has long served as a substrate for document understanding research because it provides both PDFs and machine-readable Journal Article Tag Suite (JATS) XML. 
Most prior datasets leverage this pairing by aligning text and regions extracted from PDFs to the XML through heuristic or semi-supervised matching.

\emph{GROTOAP}~\cite{tkaczyk2012grotoap} and \emph{GROTOAP2}~\cite{tkaczyk2014grotoap2} provided early large-scale ground truth for PMCOA. 
\emph{GROTOAP2} distributes hierarchical XML annotations--pages decomposed into zones, lines, words, and characters--with two-point bounding boxes and 22 zone labels (e.g., title, abstract, body, references). 

\emph{PubLayNet}~\cite{zhong2019publaynet} scales layout supervision by aligning PMCOA XML with PDFMiner~\cite{pdfminer_six} output to produce $\sim$3.5M region annotations for $\sim$360K pages. 
Regions are mapped to five canonical classes, and splits are constructed at the journal level with stricter selection for validation and test, including sampling rules to limit overrepresentation by any single journal.

Complementary to layout segmentation, \emph{PubTables-1M}~\cite{smock2022pubtables} targets table understanding: 575K pages and 948K tables annotated at the table, row, column, and cell levels, with bounding boxes in both PDF and image coordinates and word boxes provided for downstream parsing.

In contrast, our corpus is OCR-native: we bypass PMCOA XML entirely and derive word-, line-, and paragraph-level supervision directly from page images using a high-quality OCR engine, thereby avoiding alignment errors inherited from PDF parsers and enabling OCR on non-digital pages (i.e., pages containing scans without text overlays within the document).

\subsection{General-Domain OCR and Layout Resources}
\label{sec:gdomain}

\emph{IIT\mbox{-}CDIP}~\cite{lewis2006building,soboroff_cdip_2022} aggregates $\sim$7M tobacco-litigation documents (TIFF scans + text) hosted by UCSF IDL, with substantial real-world noise (handwriting, stains, scanning artifacts). 
Crucially, this text is already linearized, lacking bounding boxes for words or lines.
Subsequent work overlays OCR and structure on IIT\mbox{-}CDIP subsets. 
For example, DESSURT~\cite{davis2022end} released Tesseract~\cite{kay2007tesseract} outputs (words/lines) plus block/paragraph regions derived via PubLayNet/PrimaNet~\cite{davis_2022_6540454}. 
Widely used benchmarks curated from this source include \emph{RVL\mbox{-}CDIP} (document classification)~\cite{harley2015icdar}, \emph{FUNSD} (form understanding)~\cite{jaume2019funsd}, \emph{Tobacco\mbox{-}3482}/\emph{Tobacco\mbox{-}800} (classification, page-stream segmentation)~\cite{zhu2007automatic,lewis2006building,agam2006cdip,ucsf2007ltdl}, and \emph{DocVQA}~\cite{mathew2021docvqa}.

\emph{OCR\mbox{-}IDL}~\cite{biten2022ocr} extends this lineage by providing large-scale OCR annotations over the UCSF Industry Documents Library using a commercial engine (Amazon Textract). 
The release covers $>$26M pages (a sampled subset of a library exceeding 70M documents), enabling evaluation of systems that depend on commercial-grade OCR without bundling proprietary models.

\emph{TabMe++}~\cite{heidenreichLargeLanguageModels2024} reprocesses the \emph{TabMe} page-stream segmentation benchmark~\cite{mungmeeprued2022tab} with Azure OCR, replacing noisier Tesseract outputs. 
Although nested within the \emph{IIT\mbox{-}CDIP/IDL} universe, \emph{TabMe++} illustrates the impact of higher-quality OCR on downstream segmentation and classification.

Several resources target document layout but do not provide OCR. 
\emph{DocBank}~\cite{li2020docbank} aligns \LaTeX{} to PDF to yield 500K pages with token-level labels and PDF-derived word boxes. 
\emph{DocLayNet}~\cite{pfitzmann2022doclaynet} contributes 81K manually labeled pages ``in the wild'' across eleven region classes; earlier datasets include \emph{Marmot}~\cite{Fang2012Marmot} and the \emph{PRImA} layout benchmark~\cite{antonacopoulos2009realistic}. 
In the same parser-derived family, \emph{PDFA (PDF Association dataset)}~\cite{pixparse_pdfa_eng_wds_2024} is a large-scale subset of the SafeDocs CC\mbox{-}MAIN\mbox{-}2021\mbox{-}31 crawl~\cite{safedocs_cc_main_2021_31_pdf_untruncated_2023}, providing digital-PDF words/lines, inferred reading order, and layout metadata at web\mbox{-}scale (millions of documents). 
Unlike OCR-first corpora, PDFA recovers text from PDF objects and thus largely excludes scanned/image-only pages--a side-effect inherited by any parser-based approach.

General-domain corpora demonstrate the utility of large-scale grounded OCR and layout supervision for downstream tasks; however, scientific articles pose distinct challenges (dense mathematics, fine-grained references, heavy table/figure usage). 
\emph{PubMed\mbox{-}OCR} addresses this gap with OCR-first supervision on PMCOA pages.

\begin{table}[htb]
  \centering
  \footnotesize
  \setlength{\tabcolsep}{3.5pt}
  \renewcommand{\arraystretch}{1.08}
  \begin{tabular}{@{}>{\raggedright\arraybackslash}p{0.20\linewidth}
                  >{\raggedright\arraybackslash}p{0.12\linewidth}
                  >{\raggedright\arraybackslash}p{0.22\linewidth}
                  r r c c c c@{}}
    \toprule
    \textbf{Dataset} & \textbf{Domain} & \textbf{Engine} &
      \multicolumn{2}{c}{\textbf{Sizes}} &
      \multicolumn{4}{c}{\textbf{Granularity (bboxes)}}\\
     & & & \textbf{Docs} & \textbf{Pages} & \textbf{Blocks} & \textbf{Lines} & \textbf{Words} & \textbf{Chars} \\
    \midrule
    \multirow{2}{*}{\emph{IIT\mbox{-}CDIP}} & \multirow{2}{*}{UCSF IDL} & -- & 6.5M & 35.5M & \xmark & \xmark & \xmark & \xmark \\
    & & Tesseract & -- & 825.1K & \cmark & \cmark & \cmark & \cmark \\
    \emph{OCR\mbox{-}IDL} & UCSF IDL & Amazon Textract\textsuperscript{$\dagger$} & 4.6M & 26M & \xmark & \cmark & \cmark & \xmark \\
    \emph{TabMe++} & UCSF IDL & Azure OCR\textsuperscript{$\dagger$} & 44.8K & 122.5K & \xmark & \cmark & \cmark & \xmark \\
    \emph{PDFA} & CC & PDF Parsing (no OCR) & 2.2M & 18M & \xmark & \cmark & \cmark & \xmark \\
    \emph{GROTOAP2} & PMCOA & PDF Parsing (no OCR) & 13.2K & 119.3K & \cmark & \cmark & \cmark & \cmark \\
    \emph{PubTables\mbox{-}1M} & PMCOA & PDF Parsing (no OCR) & -- & 575.3K & \xmark & \xmark & \cmark & \xmark \\
    \textbf{\emph{PubMed\mbox{-}OCR} (Ours)} & PMCOA & Google Vision OCR\textsuperscript{$\dagger$} & \textbf{209.5K} & \textbf{1.5M} & \cmark & \cmark & \cmark & \xmark \\
    \bottomrule
  \end{tabular}
  \caption{Comparison of text resources by size and annotation granularity. Commercial engines are marked with \textsuperscript{$\dagger$}.}
  \label{tab:data-comparison}
\end{table}

\subsection{Text Recognition versus OCR}

We distinguish plain text recognition--which serializes a page into a single sequence in reading order--from grounded/structured OCR, which yields words, lines, and paragraphs with bounding boxes. 
The former is useful for ingesting documents into text corpora for LLM pre-training~\cite{kydlicek2025finepdfs,poznanskiOlmOCRUnlockingTrillions}, while grounded outputs preserve spatial provenance and enable layout-aware modeling and verifiable attribution.

Layout-aware models explicitly consume text and layout to improve document understanding, either with image encoders or by encoding layout tokens alongside text. 
Frequently, layout information is ingested in the form of bounding boxes from OCR.
Representative approaches include LayoutLM~\cite{xu2020layoutlm,huang2022layoutlmv3} and LiLT~\cite{wang2022lilt}, and more recent LLM-centric methods such as LayoutLLM (layout instruction tuning)~\cite{luo2024layoutllm}, DocLLM (LLM with layout tokens only)~\cite{wang2024docllm}, and LayTextLLM (interleaving bounding-box tokens with text)~\cite{lu2024bounding}, which demonstrate strong results without heavy vision backbones.

Grounded outputs also support \emph{grounded response generation}: answers are produced together with fine-grained evidence (citations and, when available, coordinates on the page).
Recent work on attributed/grounded generation improves verifiability by learning to attach citations at span-level granularity~\cite{huang2024learning} and by evaluating citation quality~\cite{xu2025citeeval}, with complementary advances in grounded reasoning that interleave text with bounding-box coordinates~\cite{fan2025grit}.

Because grounded OCR can be deterministically linearized when needed, it is the more verbose yet more flexible annotation. 
Our corpus therefore adopts the grounded setting with supervision at the word, line, and paragraph levels.

\section{PubMed-OCR Dataset}

\subsection{Data Collection}

We downloaded PMCOA PDFs via the official FTP/OAI endpoints and restricted redistribution to articles whose licenses permit sharing derivative artifacts. From $\sim$2M PDFs, $\sim$60\% met this criterion ($\sim$1.2M). We sample 209.5k documents uniformly at random and annotate each page with the Google Vision API (December 19, 2024 release), priced at \$1.50 per 1000 pages.
This amounts to a cost of $\sim$\$2.3k (with the cost of full OCR at roughly 5x, or \$12k).
We include only articles whose PMCOA licenses permit redistribution. For each document we release OCR JSON (always) and, where permitted, the original PDF. The metadata CSV records the license (e.g., CC BY, CC BY‑SA, CC BY‑NC, CC BY‑NC‑SA), a direct PMCID/PMID link, and allowed use (e.g., commercial use = true/false). OCR annotations are licensed under the same terms as the source article.

\subsection{OCR Processing and Normalization}

We render each PDF page to an image at 150 DPI and run Google Cloud Vision's \texttt{document\_text\_detection} on the image bytes. No manual deskewing is performed prior to calling the API with page images. From the resulting \texttt{full\_text\_annotation}, we traverse pages $\rightarrow$ blocks $\rightarrow$ paragraphs $\rightarrow$ words, extracting each word's text and its four-vertex polygon. Vertices are canonicalized to axis-aligned bounding boxes by the \{top-left, bottom-right\}. Paragraph text is formed by concatenating its words; the paragraph bounding box is the axis-aligned rectangle spanning all word vertices.

\paragraph{Line reconstruction.}
The Google Vision API only returns bounding boxes for words and paragraphs.
We derive lines by clustering words that are vertically aligned with a coarse heuristic:
\begin{enumerate}
  \item For each word $w$, let $y_{\min}(w)$ and $y_{\max}(w)$ be the minimum and maximum $y$ of its vertices, and $x_{\min}(w)$, $x_{\max}(w)$ the min/max $x$.
  \item Maintain line groups with representative $(\bar{y}_{\min}, \bar{y}_{\max})$. A word joins an existing group iff $|y_{\min}(w) - \bar{y}_{\min}| \le 5$ and $|y_{\max}(w) - \bar{y}_{\max}| \le 5$ pixels; otherwise start a new group.
  \item To avoid cross-column or cross-paragraph merges, we split any group containing words from different paragraphs according to the paragraph indices returned in the original Google Vision OCR, so each line is contained within a single paragraph.
  \item Within each group, sort words by $x_{\min}(w)$ (left-to-right) and concatenate to form the line text. The line bounding box is $\big[\min_w x_{\min}(w),\, \min_w y_{\min}(w),\, \max_w x_{\max}(w),\, \max_w y_{\max}(w)\big]$.
\end{enumerate}

\paragraph{Standardized output.}
For each page, we emit two standard artifacts: a JSON and the raw PDF. Each page JSON contains 
\texttt{text.words}, \texttt{text.lines}, and \texttt{text.paragraphs}, where each item's polygon is converted to an axis-aligned box $[X_1,Y_1,X_3,Y_3]$ (top-left, bottom-right). We also include basic image metadata (path, width, height, dpi) used to produce the OCR for reproducibility.

\subsection{Data Statistics}

\begin{table}[htb]
  \centering
  \footnotesize
  \setlength{\tabcolsep}{6pt}
  \renewcommand{\arraystretch}{1.08}
  \begin{tabular}{@{}l r r r | r r r@{}}
    \toprule
    & \multicolumn{3}{c}{\emph{PubMed\mbox{-}OCR}} & \multicolumn{3}{c}{\emph{OCR\mbox{-}IDL}} \\
    & \textbf{Count} & \textbf{Per Doc} & \textbf{Per Page} & \textbf{Count} & \textbf{Per Doc} & \textbf{Per Page} \\
    \cmidrule{2-4} \cmidrule{5-7}
    Documents & 209.5K & - & - & 4.6M & - & - \\
    Pages &  1.5M & 7.4 & - & 26M & 6 & - \\
    Paragraphs & 61.0M & 291.3 & 39.5 & - & - & - \\
    Lines & 164.3M & 784.9 & 106.3 & 46M & 101.3 & 17.5 \\
    Words & 1.3B & 6229.6 & 844.0 & 166M  & 360.8 & 62.5 \\
    \bottomrule
  \end{tabular}
  \caption{\emph{PubMed\mbox{-}OCR} corpus statistics (left) versus reported statistics from \emph{OCR-IDL} (right). We report \emph{OCR-IDL} statistics as published, but note that the number of documents/pages and their per document/page statistics imply an order of magnitude more words and lines than the manuscript purports.}
  \label{tab:ours-summary}
\end{table}

\paragraph{Comparison with prior corpora.}
Table~\ref{tab:data-comparison} situates \emph{PubMed\mbox{-}OCR} among widely used document resources. \emph{IIT\mbox{-}CDIP} is the largest in absolute size but, in its native form, lacks bounding boxes altogether; overlays such as the Tesseract pass add boxes only for a 825K\mbox{-}page subset~\cite{davis2022end}. \emph{OCR\mbox{-}IDL} and \emph{TabMe++} demonstrate the value of commercial OCR at scale in the UCSF IDL domain but omit paragraph- or character-level boxes. Parser-derived PMCOA datasets (\emph{GROTOAP2}, \emph{PubTables\mbox{-}1M}, \emph{PDFA}) recover text/regions from digital PDFs rather than page images, a process prone to reduced recall for non-digital documents. In contrast, \emph{PubMed\mbox{-}OCR} is OCR\mbox{-}first on PMCOA and provides paragraph-, line-, and word-level boxes, filling a gap between parser-derived PMCOA resources and OCR-first corpora in other domains.

\begin{figure}[htb]
    \centering
    \includegraphics[width=\linewidth]{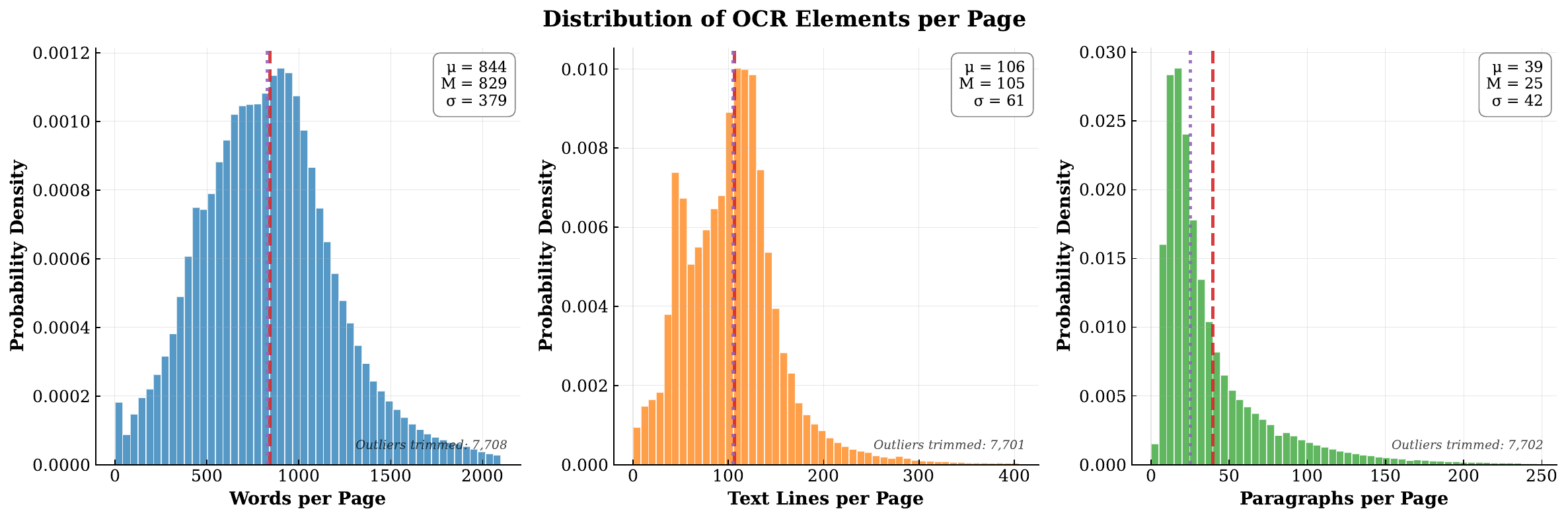}
    \caption{Distribution of number of words (left), lines (middle), and paragraphs (right) per page. $\mu$ indicates the mean, $M$ indicates the median, and $\sigma$ is the standard deviation. Each distribution is truncated at or below the 99.5th percentile to visualize the core probability mass instead of the long tail.}
    \label{fig:per_page_dist}
\end{figure}

\begin{figure}[htb]
    \centering
    \includegraphics[width=\linewidth]{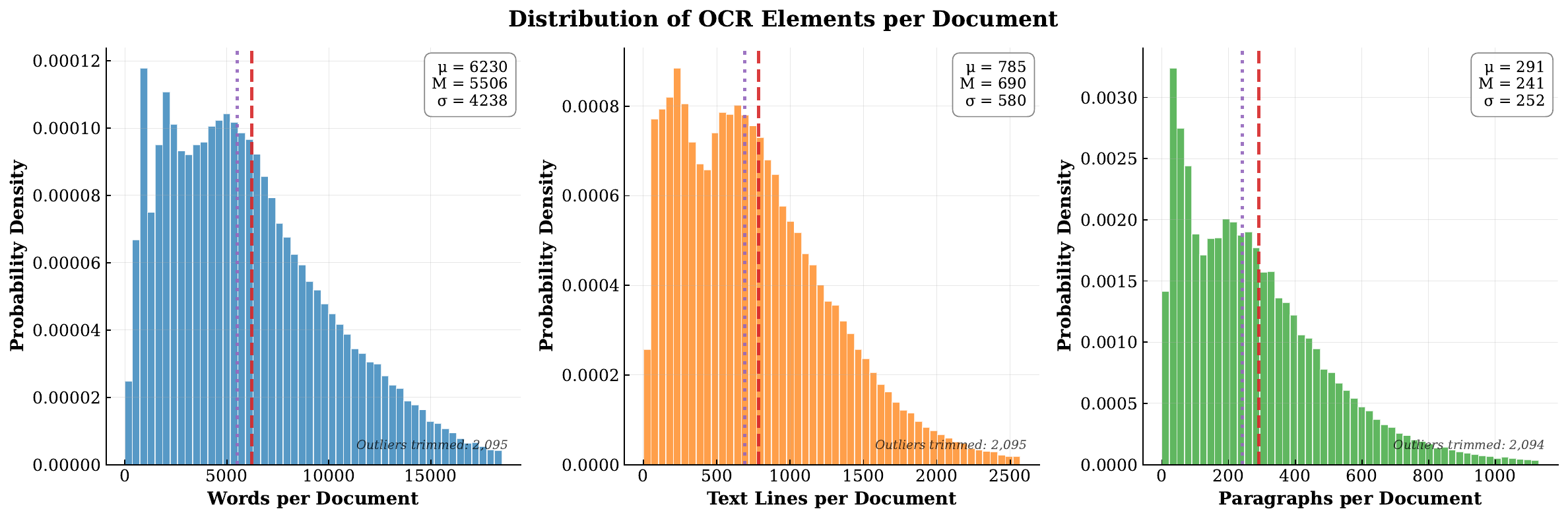}
    \caption{Distribution of number of words (left), lines (middle), and paragraphs (right) per document. $\mu$ indicates the mean, $M$ indicates the median, and $\sigma$ is the standard deviation. Each distribution is truncated at or below the 99th percentile to visualize the core probability mass instead of the long tail.}
    \label{fig:per_doc_dist}
\end{figure}

\paragraph{Corpus summary (ours).}
As shown in Table~\ref{tab:ours-summary}, the release comprises 209.5K documents and 1.5M pages (mean $7.4$ pages/doc). On average, each page contains 39.5 paragraphs, 106.3 lines, and 844 words, corresponding to 291.3 paragraphs, 784.9 lines, and 6{,}229.6 words per document. 
Comparing these statistics with the statistics reported by \emph{OCR-IDL}~\cite{biten2022ocr}, we observe that despite having fewer documents and pages, \emph{PubMed\mbox{-}OCR} has almost 4x the number of line annotations and 10x the number of word annotations.
Figures~\ref{fig:per_page_dist} and~\ref{fig:per_doc_dist} show that both per\mbox{-}page and per\mbox{-}document counts with right tails, reflecting the mix of short communications and long articles. This combination of scale and grounded granularity (paragraphs/lines/words with boxes) is designed to support layout-aware modeling, document QA with page coordinates, and robust evaluation across heterogeneous article lengths.

\begin{figure}[htb]
    \centering
    \includegraphics[width=0.9\linewidth]{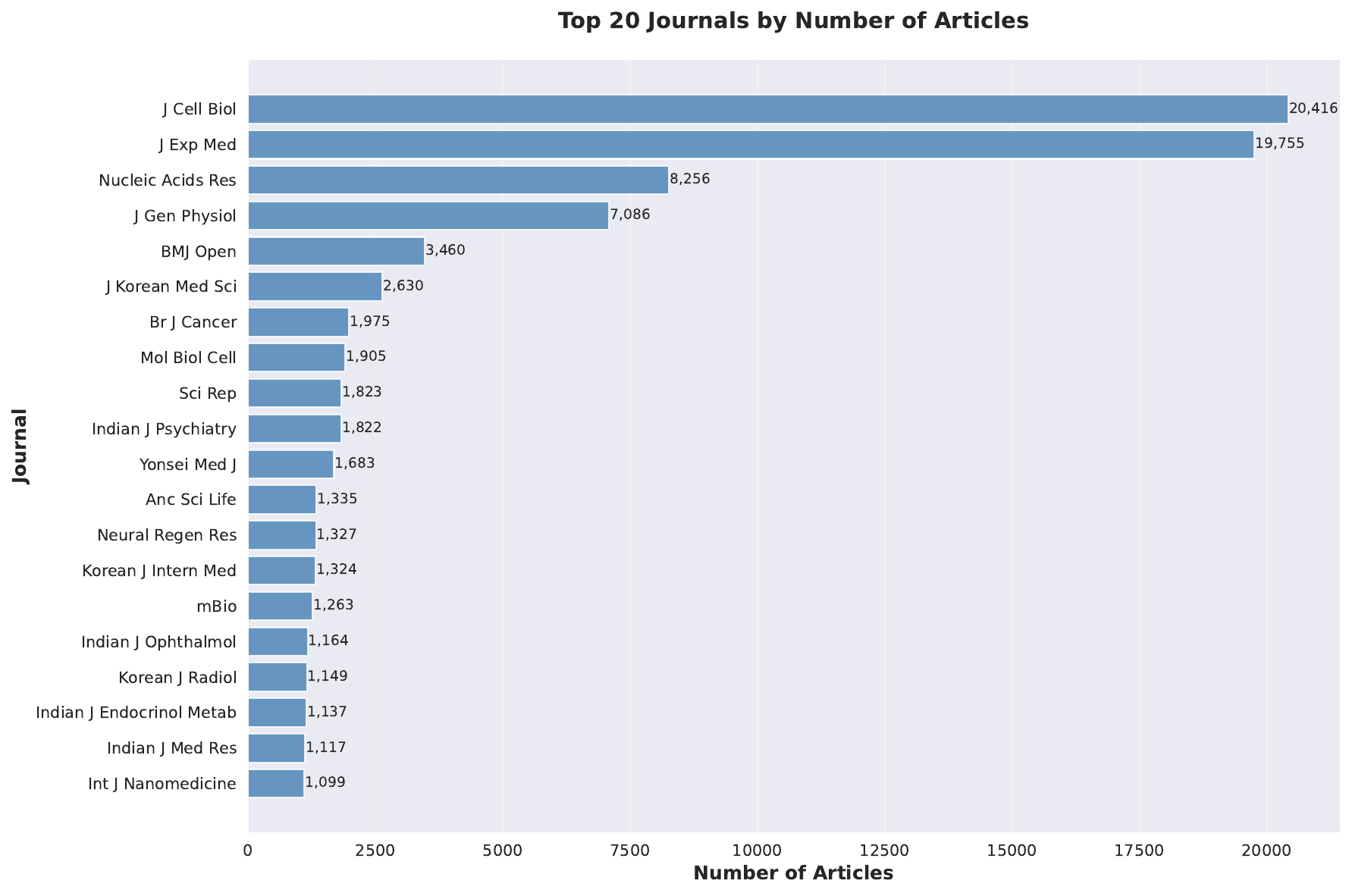}
    \caption{Top 20 journals represented in \emph{PubMed\mbox{-}OCR}. The top 3 journals account for $\sim$23\% of all documents included.}
    \label{fig:journal-dist}
\end{figure}

\paragraph{Journal distribution.}
The PMCOA composition induces a head of high\mbox{-}volume journals. The top three titles---\textit{Journal of Cell Biology} (9.7\%), \textit{Journal of Experimental Medicine} (9.4\%), and \textit{Nucleic Acids Research} (3.9\%)---account for roughly 23\% of documents. Despite this skew, 2,478 journals are represented across our dataset.
Singleton journals (journals represented with a singular document) make up 637 of the 2,478 journals, roughly 25.7\% of journals and 0.3\% of documents.
We show the top 20 journals by document count in Figure~\ref{fig:journal-dist}.

\subsection{Qualitative Analysis}

\begin{table}[htb]
    \centering
    \begin{tabular}{l r r}
    \toprule
    \textbf{Layout Feature} & \textbf{Number Detected} & \textbf{\% Pages with Feature} \\
    \midrule
    number & 36,961 & 92.09 \\
    text & 223,264 & 87.43 \\
    header & 44,349 & 79.66 \\
    footer & 36,709 & 64.96 \\
    paragraph title & 58,312 & 63.65 \\
    figure title & 28,101 & 43.99 \\
    formula & 69,480 & 24.83 \\
    image & 13,426 & 22.54 \\
    reference content & 154,944 & 19.87 \\
    reference & 13,319 & 19.80 \\
    table & 9,277 & 18.34 \\
    chart & 15,421 & 16.59 \\
    doc title & 5,286 & 12.81 \\
    footnote & 5,529 & 10.80 \\
    abstract & 4,379 & 10.32 \\
    aside text & 1,466 & 2.65 \\
    formula number & 1,238 & 1.18 \\
    algorithm & 28 & 0.06 \\
    content & 9 & 0.02 \\
    seal & 6 & 0.01 \\
    \bottomrule
    \end{tabular}
    \caption{The layout features detected across a random 40k page sample from \emph{PubMed\mbox{-}OCR}. Number detected indicates the number of each layout feature found across the entire dataset whereas the \% pages with feature indicates the percentage of pages in our sample that had at least one instance of a given layout feature. Layout features were detected using \texttt{PP-DocLayout\_plus-L}, which predicts a high prevalence of images, tables, charts, and formulas. Note that these results are model-dependent and should not be treated as gold labels.}
    \label{tab:sample-layout-features}
\end{table}

To better understand the qualitative aspects of our dataset, we sample 40,000 pages uniformly at random and run them through a pre-trained layout detection module. 
To do so, we use \texttt{PP-DocLayout\_plus-L}, which tags regions of our pages into 20 different classes. 
Some of the more interesting features include formulas (present in $\sim$25\% of pages), images (present in $\sim$22\% of pages), and charts and tables (present in $\sim$16\% and $\sim$18\%, respectively). We present the breakdown of the 20 classes of layout features in Table~\ref{tab:sample-layout-features}.

\begin{figure}[htb]
    \centering
    \includegraphics[width=0.49\linewidth]{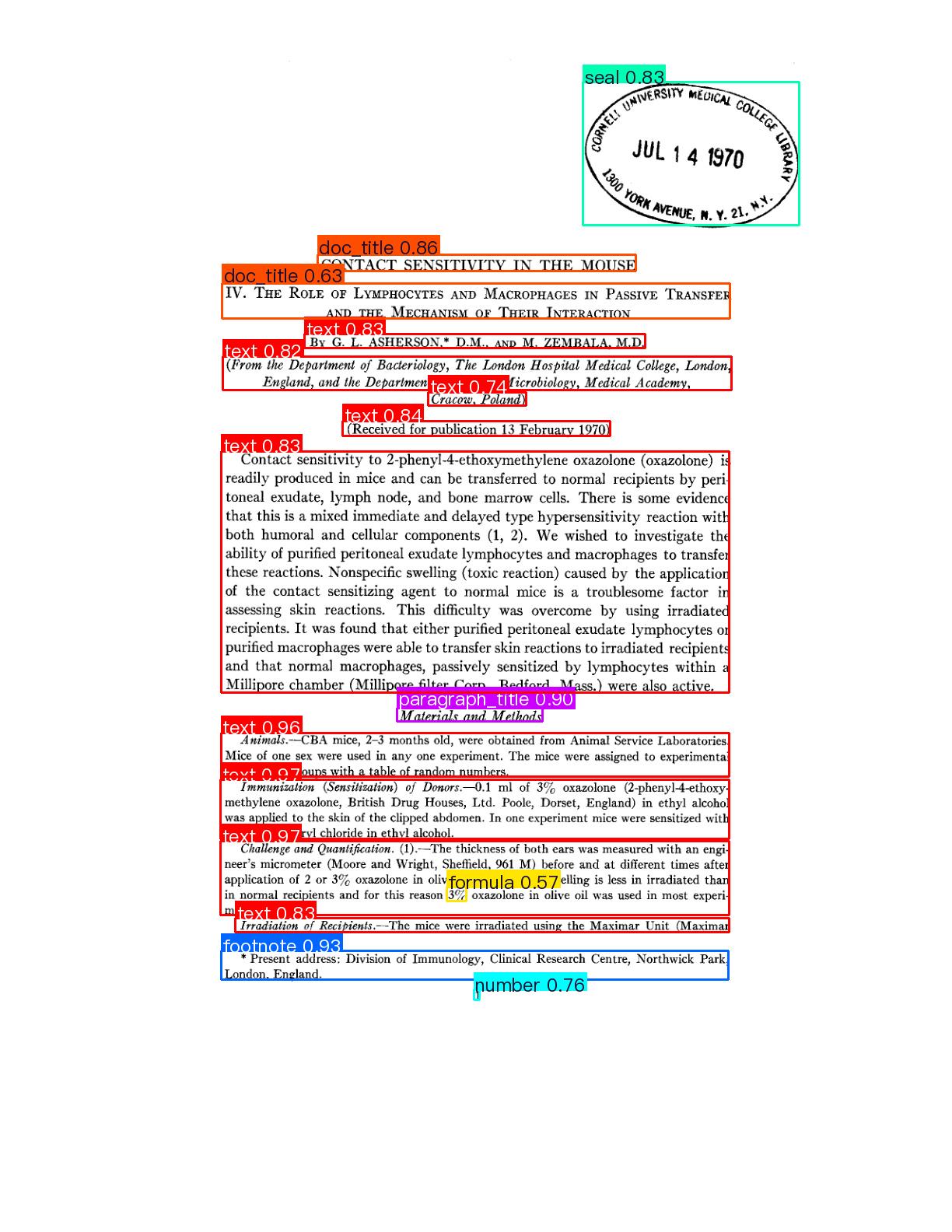}
    \includegraphics[width=0.49\linewidth]{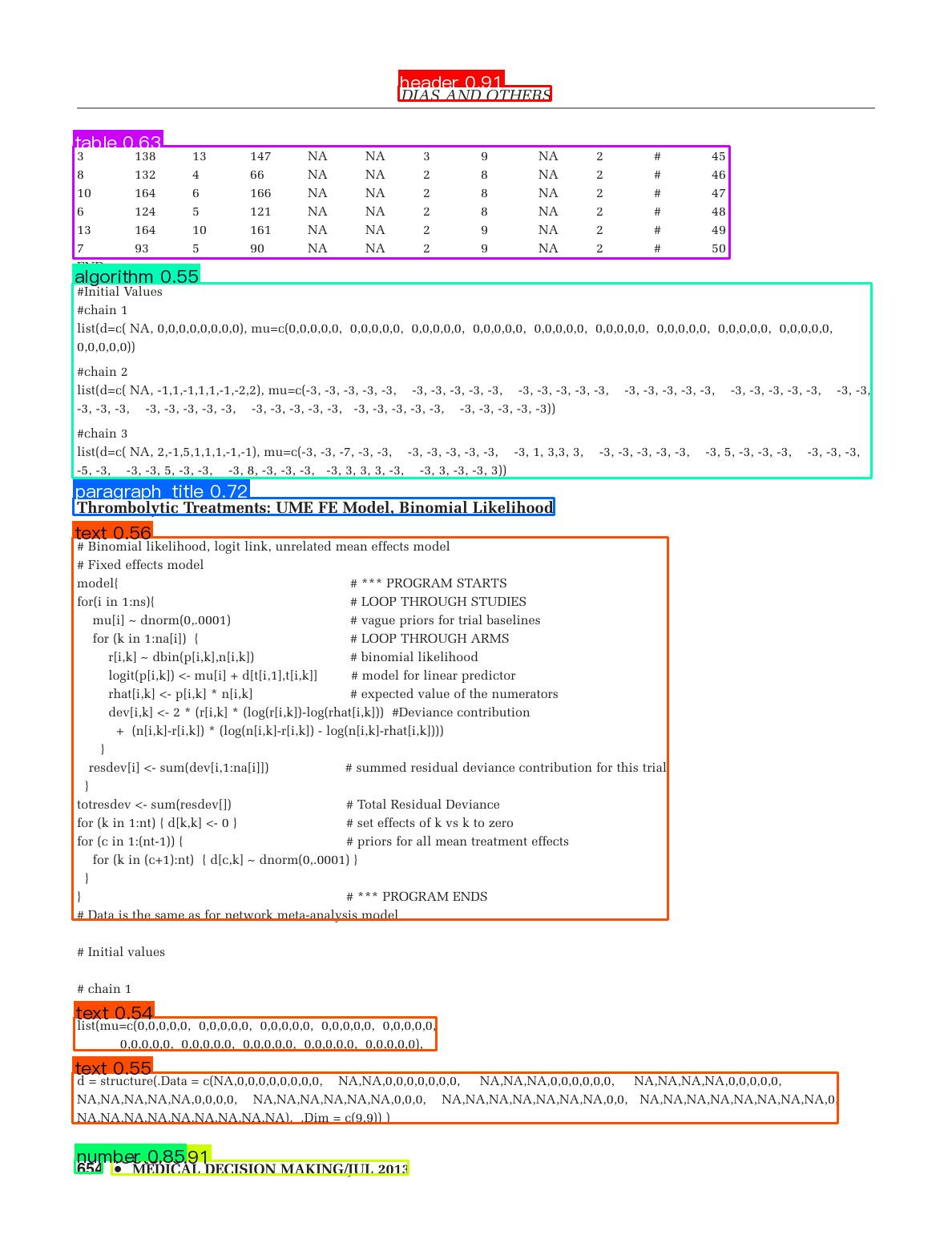}
    \caption{Two example pages from \emph{PubMed\mbox{-}OCR}, overlaid with layout detection classes predicted by PP-DocLayout. On the left, we have a page with a seal alongside high-density text (with formulas embedded within the text). On the right, we have a page with many tabular outputs, code snippets, and other text.}
    \label{fig:sample-pages-with-layout}
\end{figure}

As qualitative examples, we show two pages in Figure~\ref{fig:sample-pages-with-layout}. The first is an older document with a stamped seal on its upper-right. It has a mixture of dense text features and formulas embedded within that text. The second image shows tabular data, algorithmic definitions, among other standard text and title features.
We provide a handful of additional samples in Appendix~\ref{app:samples}.

\section{Conclusion}
We presented \emph{PubMed\mbox{-}OCR}, an OCR-first corpus derived from the PubMed Central Open Access subset that exposes paragraph-, line-, and word-level bounding boxes directly from page images in a compact, standardized JSON format. By bypassing fragile PDF/XML alignment, the resource complements parser-based PMCOA derivatives and enables layout-aware modeling, grounded question answering, and attributed generation on scientific literature. We also provide corpus-level statistics and distributions to characterize scale and diversity across journals and article lengths, making the dataset a practical substrate for training and evaluation.

While useful, the corpus has limitations. It currently relies on a single OCR engine, and line annotations are reconstructed heuristically from word boxes, which may introduce biases in reading order and grouping. Character-level boxes and explicit representations of mathematical expressions or figure/table structure are not included, and coverage reflects PMCOA’s license and journal distribution. These constraints should be considered when reporting results and designing experimental splits.

Taken together, \emph{PubMed\mbox{-}OCR} offers a reproducible, openly accessible dataset for research that requires faithful text–layout grounding in scientific articles. We release the resource to support robust evaluation and to facilitate fair comparisons without dependence on proprietary pipelines, and we invite the community to audit, extend, and build upon it.

\bibliographystyle{unsrt}  
\bibliography{references}  

\appendix

\section{More Examples}
\label{app:samples}

\begin{figure}
    \centering
    \includegraphics[width=0.6\linewidth]{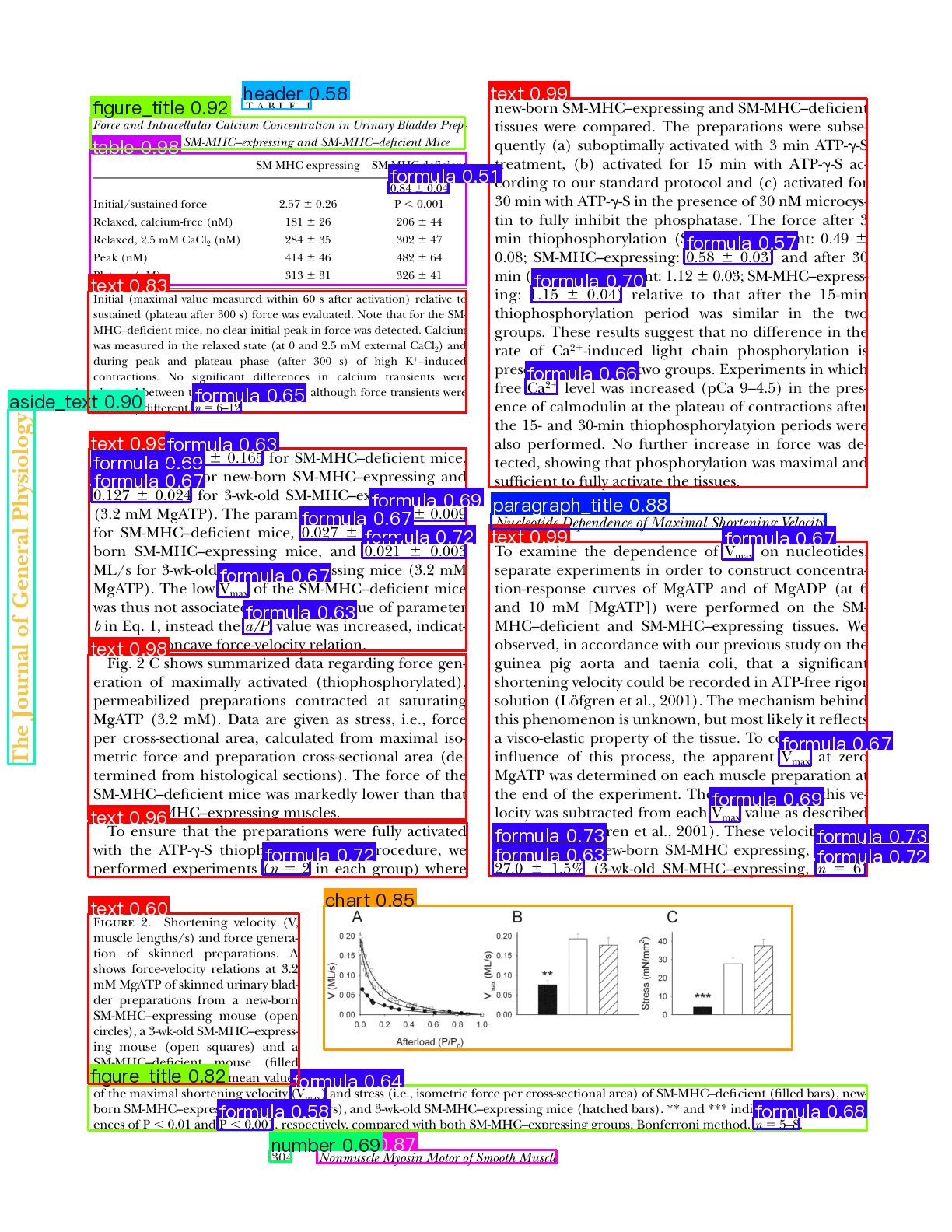}
    \caption{A sample page from \emph{PubMed\mbox{-}OCR} exhibiting a variety of features: aside text, charts, captions, and formulas.}
    \label{fig:sample1}
\end{figure}

\begin{figure}
    \centering
    \includegraphics[width=0.8\linewidth]{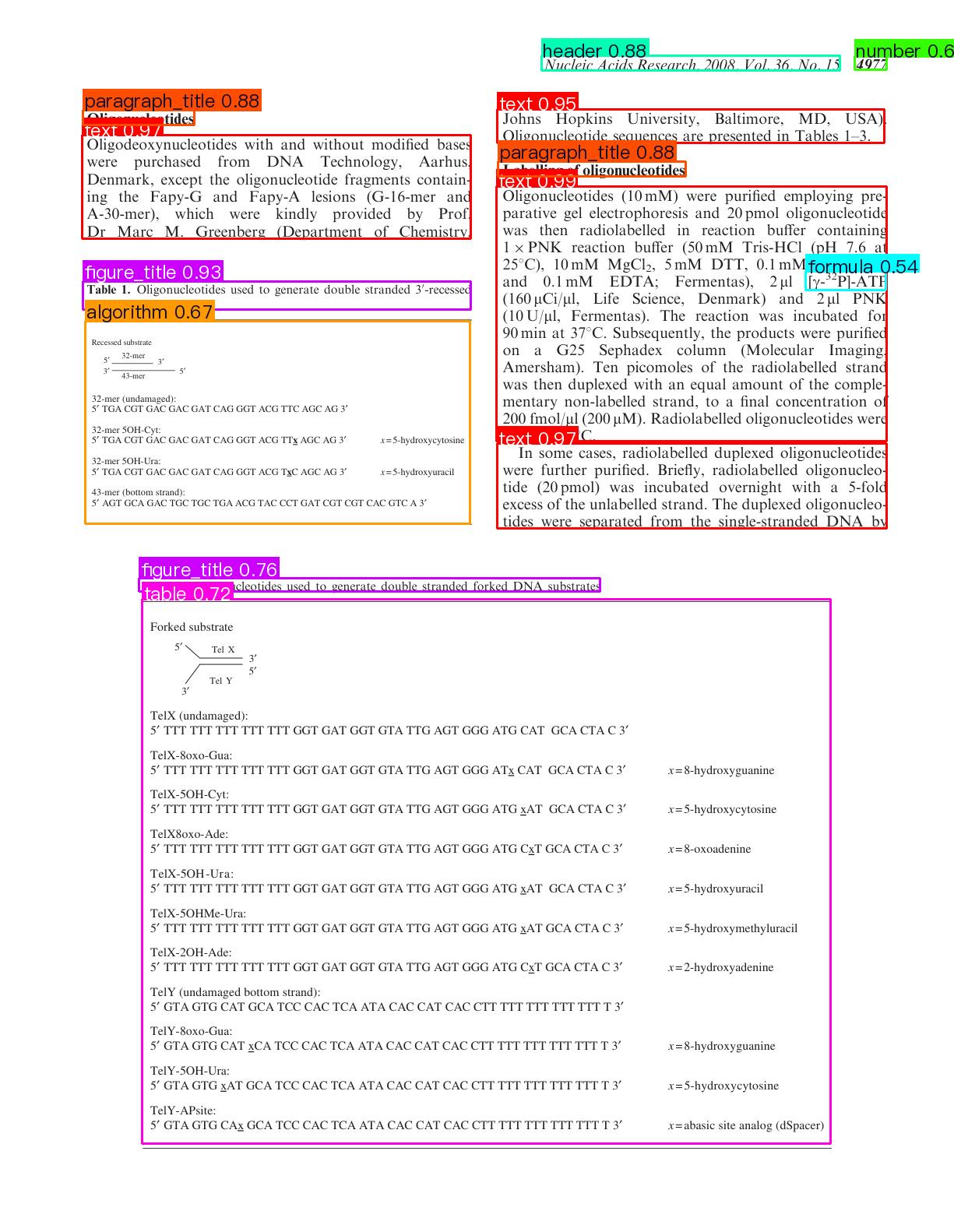}
    \caption{A sample with a complex scientific table. A second tabular section is mis-identified as an algorithm. Introducing \emph{PubMed\mbox{-}OCR} with layout annotations could be a valuable augmentation that would benefit current generation layout models.}
    \label{fig:sample2}
\end{figure}

\begin{figure}
    \centering
    \includegraphics[width=0.8\linewidth]{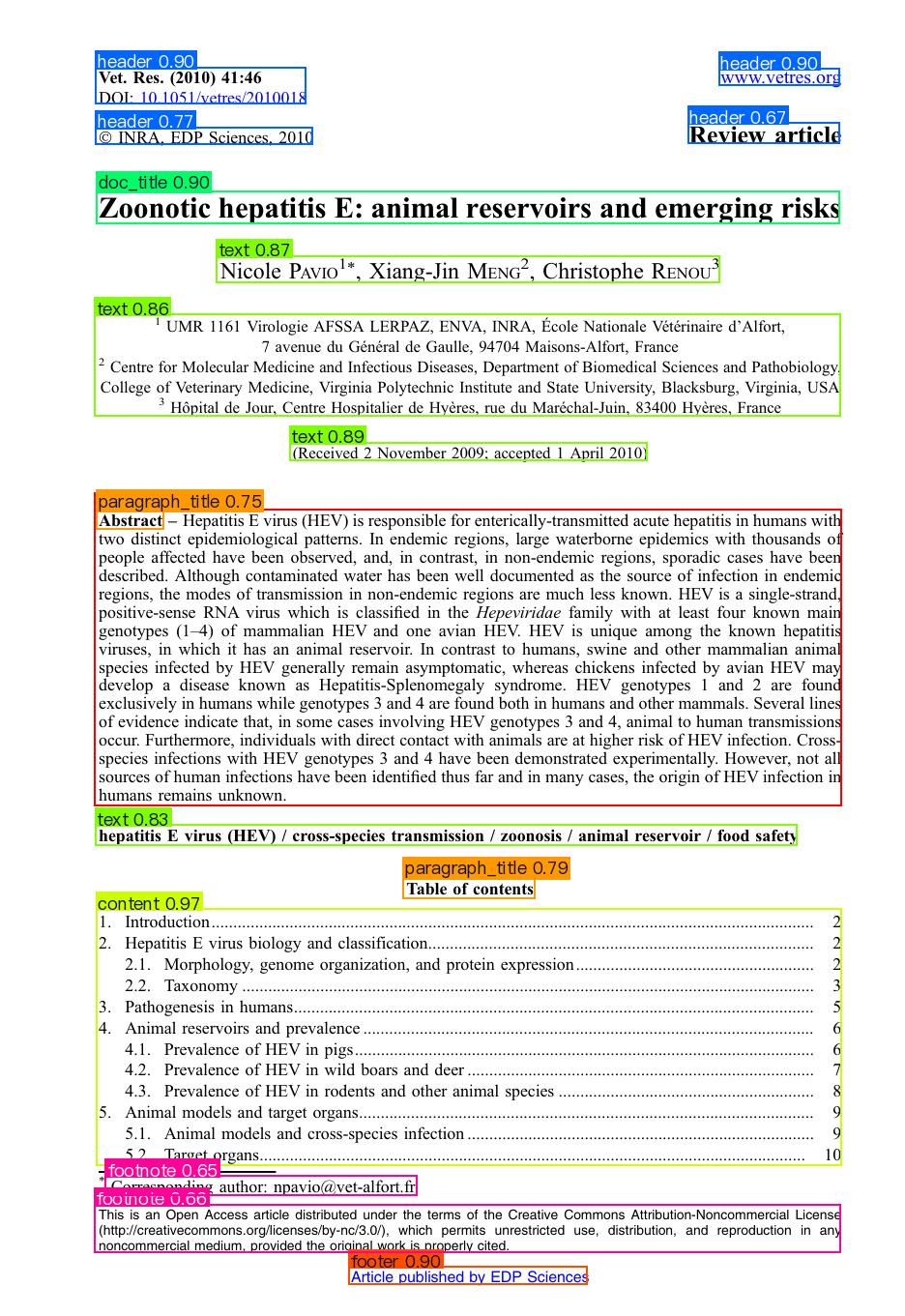}
    \caption{A sample page with dense text and a table of contents as well as structured detection of paragraph and document titles.}
    \label{fig:sample3}
\end{figure}

\begin{figure}
    \centering
    \includegraphics[width=0.8\linewidth]{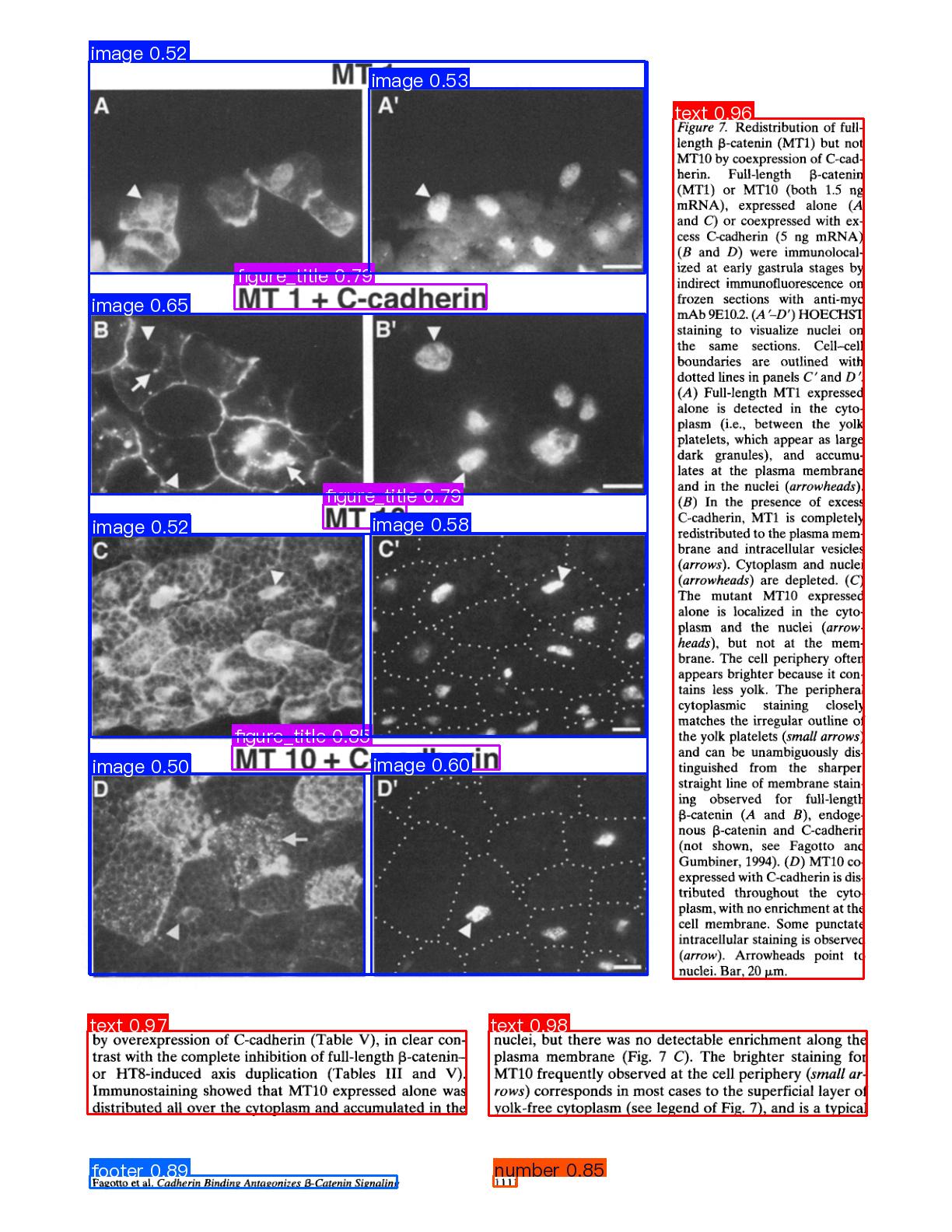}
    \caption{A sample page that is image-dense with structured captions about the details and intended interpretation.}
    \label{fig:sample4}
\end{figure}

We show a handful of additional samples with layout detection annotations from PP-DocLayout in Figures~\ref{fig:sample1}--~\ref{fig:sample4}.

\end{document}